\pdfoutput=1

\documentclass[11pt]{article}

\usepackage[]{acl}

\usepackage{times}
\usepackage{latexsym}

\usepackage[T1]{fontenc}

\usepackage[utf8]{inputenc}

\usepackage{microtype}

\usepackage{inconsolata}
\usepackage[graphicx]{realboxes}
\usepackage{ragged2e} 
\usepackage{makecell}
\usepackage{float}
\usepackage{stfloats}

\usepackage{xurl}
\usepackage{multirow}

\title{"Vorbești Românește?" \\ A Recipe to Train Powerful Romanian LLMs with English Instructions}

\author{ 
\bf Mihai Masala$^{a,b,c}$, Denis C. Ilie-Ablachim$^b$, Alexandru Dima$^{a,b}$, Dragos Corlatescu$^b$,\\ \bf Miruna Zavelca$^{a,d}$,
 Ovio Olaru$^f$, Simina Terian$^f$, Andrei Terian$^f$, Marius Leordeanu$^{b,c}$,\\ \bf Horia Velicu$^e$, Marius Popescu$^d$,
 Mihai Dascalu$^b$, Traian Rebedea$^{b,g}$\\
$^a$Institute for Logic and Data Science, Bucharest,\\ $^b$National University of Science and Technology POLITEHNICA Bucharest, Romania,\\ $^c$Simion Stoilow Institute of Mathematics of The Romanian Academy, Bucharest,\\ $^d$University of Bucharest, Romania, $^e$BRD Groupe Societe Generale,\\ $^f$Lucian Blaga University Sibiu, $^g$NVIDIA
}
\date{}

\begin{document}
\maketitle
\begin{abstract}
In recent years, Large Language Models (LLMs) have achieved almost human-like performance on various tasks. While some LLMs have been trained on multilingual data, most of the training data is in English; hence, their performance in English greatly exceeds other languages. To our knowledge, we are the first to collect and translate a large collection of texts, instructions, and benchmarks and train, evaluate, and release open-source LLMs tailored for Romanian. We evaluate our methods on four different categories, including academic benchmarks, MT-Bench (manually translated), and a professionally built historical, cultural, and social benchmark adapted to Romanian. We argue for the usefulness and high performance of RoLLMs by obtaining state-of-the-art results across the board. We publicly release all resources (i.e., data, training and evaluation code, models) to support and encourage research on Romanian LLMs while concurrently creating a generalizable recipe, adequate for other low or less-resourced languages.


\end{abstract}

\section{Introduction}
\label{sec:introduction}

The Transformer architecture~\cite{vaswani2017attention} has become ubiquitous recently in the Natural Language Processing (NLP) domain, leading to state-of-the-art performance on various tasks, ranging from text classification to text generation. Decoder-only language models such as GPT series~\cite{radford2018improving,radford2019language,brown2020language} exhibit impressive capabilities such as understanding and generating natural language, learning and solving tasks not directly trained on. In the last few years, Large Language Models (LLMs) development has exploded, with a plethora of models available~\cite{achiam2023gpt,chowdhery2023palm,touvron2023llama,hoffmann2022training,alpaca,jiang2023mistral}.

Training such massive models requires a vast amount of available data (ranging from tens to hundreds of billions of text tokens) besides computational resources. Intrinsically, most work on such models focuses mainly on English, often excluding other languages. Here, we focus on building the resources for training LLMs specialized in Romanian, a less-resourced language. By showcasing our approach and releasing our models, we strive to enable further research and exciting applications for Romanian-speaking users.

Our contributions can be summarized as follows:
\textbf{I.} We present and release the first Romanian LLMs with both foundational and instruct variants, based on popular open-source models such as Llama2~\cite{touvron2023llama}, Llama3\footnote{\url{ai.meta.com/blog/meta-llama-3/}}, Mistral~\cite{jiang2023mistral} and Gemma~\cite{team2024gemma}. Our collection of models is grouped under the OpenLLM-Ro initiative\footnote{\url{www.openllm.ro}}; \textbf{II.} We collect and translate a large set of evaluation benchmarks (including a human translation of MT-Bench~\cite{zheng2024judging}), conversation, and instructions; \textbf{III.} We extend and adapt existing testing suites to thoroughly assess the performance of different models on Romanian tasks. \textbf{IV.} We propose a novel methodology to assess language-specific LLMs by proposing the first benchmark that evaluates the Romanian cultural knowledge of LLMs; \textbf{V.} We make our recipe (i.e., datasets and code) and models open-source to encourage further research both for Romanian and other low/less-resourced languages.


\section{Related Work}
\label{sec:related_work}

With the advent of open-source LLMs, especially Llama2~\cite{touvron2023llama} and Mistral~\cite{jiang2023mistral}, non-English specialized alternatives have been trained. ~\citet{luukkonen2023fingpt} have trained a family of Finnish foundational models, ranging from 186M to 176B parameters. Trained on a total of 38B tokens with a custom tokenizer, their models showcase state-of-the-art performance on a variety of Finish downstream tasks. Similarly, \citet{cui2023efficient} have trained both a foundational and a chat variant of Llama2 for Chinese, while \citet{wang2023mediagpt} worked on a domain-specific (i.e., media) Chinese LLM. 
For Italian, \citet{bacciu-etal-2024-dantellm-lets} translated four out of the six benchmarks used in Open LLM Leaderboard\footnote{\url{www.huggingface.co/spaces/HuggingFaceH4/open\_llm\_leaderboard}} to release the first Italian LLM benchmark and evaluate existing Italian models~\cite{basile2023llamantino,bacciu2023fauno,santilli2023camoscio,bacciu-etal-2024-dantellm-lets}, models that are all adapter-based fine-tunes. Other language-specific LLMs include versions for 
Dutch\footnote{\url{www.huggingface.co/BramVanroy/GEITje-7B-ultra}}, Bulgarian\footnote{\url{www.huggingface.co/INSAIT-Institute}}, Serbian~\cite{YugoGPT}, or Japanese~\cite{fujii2024continual}.

\citet{dac2023okapi} automatically translated English Alpaca~\cite{alpaca} instructions together with a set of generated Self-Instruct~~\cite{wang-etal-2023-self-instruct} instructions into 26 languages. Furthermore, they translated for evaluation three out of the six benchmarks used in the Open LLM Leaderboard. They fine-tuned Llama~\cite{touvron2023llama-1}, thus obtaining language-specific models. \citet{li2023bactrian} translated the Alpaca and Dolly~\cite{DatabricksBlog2023DollyV2} datasets into 51 languages and fine-tuned the Llama model, further evaluated in a zero-shot setup on downstream tasks.

\citet{singh2024aya} contributed with a huge amount of multilingual text resources under the Aya initiative.
Over 2000 people across 119 countries contributed to the creation of Aya Dataset, a dataset of over 204k human-curated instructions in 65 languages. The Aya Collection uses instruction-style templates created by fluent speakers and applies them to existing datasets, generating over 500M instances that cover 114 languages. Based on Aya resources, \citet{ustun2024aya} fine-tuned Aya-101, a 13B parameter mT5 model~\cite{xue-etal-2021-mt5} capable of following instructions in 101 languages. Compared to Aya-101, which focuses on breadth (i.e., a lot of languages), ~\citet{aryabumi2024aya} considered a more depth-based approach, building Aya-23, a 23-language multilingual instruct model following Cohere’s Command model\footnote{\url{www.cohere.com/command}}.

Based on previous research~\cite{yong-etal-2023-bloom,zhao2024llama}
, we decided to perform both continual pre-training and instruction fine-tuning.

\section{Data}
\label{sec:data}

We employ two distinct data sources for training a Romanian LLM: documents for training the foundational model and instructions for the chat variant. 

\subsection{CulturaX}

CulturaX~\cite{nguyen2023culturax} is a collection of documents in 167 languages, built on top of mC4~\cite{xue-etal-2021-mt5} and OSCAR~\cite{abadji-etal-2022-towards} and using multiple cleaning and deduplication stages to ensure a high-quality corpus. Around 40M of these documents are in Romanian amounting to 40B tokens.

\subsection{Instructions and Conversations}


We collect a large and diverse dataset of conversations and instructions to build the instruct models. Since high-quality data in Romanian is very scarce, we resort to translating\footnote{\url{www.systransoft.com/} last accessed 6th of March 2024} datasets such as Alpaca~\cite{alpaca}, Dolly~\cite{DatabricksBlog2023DollyV2}, NoRobots~\cite{no_robots}, GPT-Instruct\footnote{\url{www.github.com/teknium1/GPTeacher}}, Orca~\cite{mukherjee2023orca}, Camel~\cite{li2023camel}, and UltraChat~\cite{ding2023enhancing}. Furthermore, we use Romanian data from OpenAssistant~\cite{kopf2023openassistant} and Aya Collection~\cite{singh2024aya}. This led to a total of around 2.7M instructions and conversations in Romanian, which were further used in our experiments.

\section{Training}
\label{sec:training}

Initial pre-training experiments were limited to a maximum sequence length of 512 to accelerate training. We further experimented with longer sequence lengths (i.e., 1024 and 2048), always picking the biggest batch size that fits into the memory. 
For all experiments, we used an AdamW optimizer with weight decay set to 0.1 and gradient clipping. For both continual pre-training and instruction fine-tuning, we performed full training (i.e., updating all model parameters).

\subsection{Foundational Model}

Training the foundational model is done iteratively using continual pre-training on the CulturaX dataset. Starting from the foundational Llama2 model, we trained the first model on 5\% of the CulturaX, the following model on the next 5\% (leading to the second model seeing 10\% of the data), and the third model on another 10\% of the CulturaX. Therefore, we have three variants of a Romanian foundational model, variants that used 5\%, 10\%, and 20\% of the entire CulturaX dataset. Training is done over one epoch, with a fixed learning rate of $1e^{-4}$. 

\subsection{Instruct Models}

RoLlama2-Instruct is built by fine-tuning the Romanian Llama2 base model, on one epoch with a fixed learning rate of $5\times 10^{-5}$. We further built Romanian instruct variants of other popular models (i.e., Mistral-7b, Gemma-7b, and Llama3-8b), dropping the pre-training and directly performing instruction fine-tuning on the base models. For these models, a learning rate of $10^{-6}$ for two epochs worked best. 

\section{Evaluation}
\label{sec:evaluation}

\begin{table*}[ht]
\centering
\begin{tabular}{lccccccc}
\hline
\textbf{Model} & \textbf{Average} & \textbf{ARC} & \textbf{MMLU} & \textbf{Wino} & \textbf{HS} &  \textbf{GSM8k} & \textbf{TQA}\\
\hline

\multicolumn{8}{c}{\textit{\textbf{Llama2}}}\\
Llama-2-7b & 37.04	& 36.05 &	33.66 &	57.56 &	48.00 &	4.75 &	42.22\\
\textit{RoLlama2-7b-Base} (512-5\%) & 38.13	&37.66	&31.56	&59.51	&56.25	&3.29	&40.51\\
\textit{RoLlama2-7b-Base} (1024-5\%) & 39.20	&37.99	&31.99	&59.71	&56.91	&4.12	&44.51\\
\textit{RoLlama2-7b-Base} (2048-5\%) & \textbf{39.31}	&38.35	&30.29	&59.30	&57.36	&5.08	&45.48\\
\textit{RoLlama2-7b-Base} (1024-10\%) & 38.42&	38.09&	30.28&	59.00&	57.58&	2.98	&42.58\\
\textit{RoLlama2-7b-Base} (1024-20\%) & 38.03&	37.95	&27.22&	59.29&	57.22	&2.53	&44.00\\

\hline
Llama-2-7b-chat & 36.84 &	37.03 &	33.81 &	55.87 &	45.36 &	4.90 &	44.09\\
\textit{RoLlama2-7b-Instruct} & \textbf{44.50} & 44.73 & 40.39 & 63.67 & 59.12 & 13.29 & 45.78 \\

\hline

\multicolumn{8}{c}{\textit{\textbf{Mistral}}}\\
Mistral-7B-v0.1  & 45.02 &	42.99 & 47.16 & 60.77 & 54.19 & 16.20 & 48.80\\
Mistral-7B-Instruct-v0.2  & 47.40 & 46.29 & 47.01 & 58.78 & 54.27 & 13.47 & \textbf{64.59}\\
\textit{RoMistral-7b-Instruct} & \textbf{52.91} & \textbf{52.27} & 49.33 & \textbf{70.03} & \textbf{62.88} & 32.42 & 50.51\\

\hline

\multicolumn{8}{c}{\textit{\textbf{Llama3}}}\\
Llama-3-8B & 44.55 & 38.05 & 48.33 & 59.94 & 53.48 & 20.04 & 47.44\\
Llama-3-8B-Instruct  & 50.62 & 43.69 & 52.04 & 59.33 & 53.19 & 43.87 & 51.59\\
\textit{RoLlama3-8b-Instruct} & \textbf{52.21} &	47.95 & 53.50 & 66.06 & 59.72 & 40.16 & 45.90\\

\hline

\multicolumn{8}{c}{\textit{\textbf{Llama3.1}}}\\
Llama-3.1-8B & 37.29 & 33.25 & 36.35 & 58.80 & 42.65 & 3.59 & 49.03\\
Llama-3.1-8B-Instruct  & 49.87 & 42.86 & 53.73 & 59.71 & 56.82 & 35.56 &50.54\\
\textit{RoLlama3.1-8b-Instruct} & \textbf{53.03} & 47.69 & \textbf{54.57} & 65.85 & 59.94 & \textbf{44.30} & 45.82\\

\hline

\multicolumn{8}{c}{\textit{\textbf{Gemma}}}\\
gemma-7b & 50.04 & 47.22 & 53.18 & 61.46 & 60.32 & 30.48 & 47.59\\
gemma-1.1-7b-it & 41.39	& 40.05&	47.12	&54.62	& 47.10	&9.73	&49.75\\
\textit{RoGemma-7b-Instruct} & \textbf{50.48} & 52.02 & 52.37 & 66.97 & 56.34 & 25.98 & 49.18\\

\hline



\multicolumn{8}{c}{\textbf{\textit{Other models}}}\\
Okapi-Ro & 35.64 & 37.90 & 27.29 & 55.51 & 48.19 & 0.83 & 44.15\\
aya-23-8B & \textbf{45.81} & 43.89 & 45.96 & 60.50 & 60.52 & 16.81 & 47.16\\
\hline
\end{tabular}
\vspace*{-0.6em}
\caption{Comparison between RoLLMs and other LLMs on Romanian versions of academic benchmarks (abbreviations: (1024-10) - sequence length of 1024, 10\% of CulturaX used for pretraining, HS - HellaSwag, Wino - Winogrande, TQA - TruthfulQA). \textbf{Bold} denotes the best in each category (average) and overall (each benchmark).}
\label{tab:overview}
\end{table*}

Evaluating a general LLM is not straightforward, considering the vast number of benchmarks and evaluation protocols proposed and evaluating LLMs in any language besides English only adds another layer of complexity. We propose an extensive set of evaluations, grouped into four different categories, to obtain a thorough performance assessment of LLMs in Romanian.

\subsection{Academic Benchmarks}
Popular benchmarks include Open LLM Leaderboard, an online leaderboard that encompasses six reasoning and general knowledge tasks: ARC~\cite{clark2018think}, HellaSwag~\cite{zellers2019hellaswag}, MMLU~\cite{hendrycks2020measuring}, TruthfulQA~\cite{lin2021truthfulqa}, Winogrande~\cite{sakaguchi2021winogrande} and GSM8k~\cite{cobbe2021training}. 

For Romanian versions of ARC, HellaSwag, MMLU, and TruthfulQA, we resort to ChatGPT translations~\cite{dac2023okapi}, while we translate\footnote{same system used for translating instructions} Winogrande and GSM8k. The generally used few-shot values for ARC, HellaSwag, and MMLU (i.e., 25 shots for ARC, 10 shots for HellaSwag, and 5 shots for MMLU) lead to the prompt having a length equal to or even surpassing 2048 tokens. As some of our models have been trained only on sequences of length 512, 1024, or 2048, we opted to use multiple few-shot settings. More specifically, for ARC, we tested with 0, 1, 3, 5, 10, and 25 shots, MMLU and Winogrande with 0, 1, 3, and 5, and HellaSwag with 0, 1, 3, 5, and 10 shots. For GSM8k, we employ 1, 3, and 5 shot evaluation. Results in Table~\ref{tab:overview} are averaged over all few shot settings for each benchmark.

\subsection{MT-Bench}
MT-Bench proposes a multi-turn instruction set denoting that strong LLMs (e.g., GPT-4) can act as judges and approximate human preferences in a scalable and somewhat explainable way. While most benchmarks in the Open LLM Leaderboard focus more on the intrinsic knowledge of a model, MT-Bench focuses more on the conversation capability and the quality of the generated answers. In our experiments, we use GPT-4o\footnote{\url{www.openai.com/index/hello-gpt-4o/} last accessed on 11$^{th}$ of June 2024} as a judge. We decided to manually translate MT-Bench because it is a small and highly relevant benchmark for instruct models and it represents a crucial clean test set for Romanian models.

\subsection{Romanian Downstream Tasks}
Besides the standard benchmarks that are used for evaluating LLMs, we further evaluate a series of Romanian downstream tasks to assess the "real-world" performance of our models in solving Romanian tasks. We selected four tasks from the LiRo benchmark~\cite{liro2021}: sentiment analysis on LaRoSeDa~\cite{tache-etal-2021-clustering}, machine translation on WMT-16-ro-en dataset~\cite{bojar2016results}, question answering on XQuAD-ro~\cite{Artetxe:etal:2019} and semantic text similarity on RO-STS~\cite{liro2021}. We framed sentiment analysis as a classification task, whereas machine translation, question answering, and semantic text similarity are framed as text generation problems. For semantic text similarity, we follow the approach of~\citet{gatto-etal-2023-text}, with further details presented in Appendix Section~\ref{sec:app_prompts}.

For all tasks, we perform zero/few-shot evaluation (0, 1, 3, and 5 shot averaged) and also fine-tune the models. We used LoRA \cite{hu2021lora} to ensure equitable comparisons between models, targeting all linear layers, with a rank and alpha of 8, a 0.1 dropout, and a learning rate of $5\times 10^{-5}$. In our initial experiments, we found this approach to generate the most consistent results. We fine-tuned the models for each dataset for only one epoch as we found that this was more than enough for the loss to stabilize.


\begin{table*}[ht]
\centering
\begin{tabular}{lccc|c}
\hline
\textbf{Model} & \textbf{Average} & \textbf{1st turn} & \textbf{2nd turn} & \textbf{\#Answers in Ro}\\
\hline

Llama-2-7b-chat & 0.59 & 0.95 & 0.23 & 21 / 160 \\
Llama-2-7b-chat (\textit{ro\_prompted}) & 1.08 & 1.44 & 0.73 & 45 / 160 \\
\textit{RoLlama2-7b-Instruct} & \textbf{4.43} & 4.93 & 3.94 & \textbf{160 / 160}\\
\hline

Mistral-7B-Instruct-v0.2 & 1.04 & 1.35 & 0.72 & 27 / 160 \\
Mistral-7B-Instruct-v0.2 (\textit{ro\_prompted}) & 5.03 & 5.05 & 5.00 & 154 / 160 \\
\textit{RoMistral-7b-Instruct} & \textbf{5.29} & 5.86 & 4.73 & \textbf{160 / 160}\\

\hline

Llama-3-8B-Instruct & 4.54 & 4.65 & 4.43 & 126 / 160\\
Llama-3-8B-Instruct (\textit{ro\_prompted}) & \textbf{5.96} & \textbf{6.16} & \textbf{5.76} & 158 / 160\\
\textit{RoLlama3-8b-Instruct} & 5.38 & 6.09 & 4.68 & \textbf{160 / 160}\\

\hline

Llama-3.1-8B-Instruct & \textbf{5.69} & 5.96 & 5.43 & 160 / 160\\
Llama-3.1-8B-Instruct (\textit{ro\_prompted}) & \textbf{5.69} & 5.85 & 5.53 & 160 / 160\\
\textit{RoLlama3.1-8b-Instruct} & 5.42 & 5.95 & 4.89 & \textbf{160 / 160}\\

\hline

gemma-1.1-7b-it & 5.09 & 5.53 & 4.66 & 158 / 160\\
gemma-1.1-7b-it (\textit{ro\_prompted}) & 4.83 & 5.11 & 4.55 & \textbf{160 / 160}\\
\textit{RoGemma-7b-Instruct} & \textbf{5.24} & 5.55 & 4.94 & \textbf{160 / 160}\\

\hline


\hline


\hline
\end{tabular}
\vspace*{-0.6em}
\caption{Romanian MT-Bench performance. \textbf{Bold} as in Table~\ref{tab:overview}.}
\label{tab:chat-mtbench}
\end{table*}

\begin{table*}[ht]
\centering
\begin{tabular}{lcc|cc}
\hline
\multirow{2}{*}{\textbf{Model}}& \multicolumn{2}{c}{\textbf{Zero/Few shot}} & \multicolumn{2}{c}{\textbf{Fine-tuned}} \\
& Binary (\textit{F1}) & Multiclass (\textit{F1}) & Binary (\textit{F1}) & Multiclass (\textit{F1})\\
\hline
SOTA - Romanian BERT$^{a}$& - & - & 98.31 & 86.63\\
\hline
Llama-2-7b & 93.19	& 54.11	& 98.43	& 87.22	\\
\textit{RoLlama2-7b-Base} & 83.25	& 61.04	&98.97	&87.72\\
\hline
Llama-2-7b-chat & 87.78	& 52.82&	97.27	&82.02	\\
\textit{RoLlama2-7b-Instruct} & \textbf{97.66}	&62.41	&97.97	&60.89\\
\hline
Mistral-7B-Instruct-v0.2 & 96.97 &	56.66	&98.83	&87.32 \\
\textit{RoMistral-7b-Instruct} & 95.56	&\textbf{67.83}	&\textbf{99.00}	&\textbf{87.57}	\\
\hline
Llama-3-8B-Instruct & 95.88&	56.21	&98.53	&86.19 \\
\textit{RoLlama3-8b-Instruct} & 95.58	&61.20	&96.46	&87.26\\
\hline
Llama-3.1-8B-Instruct & 95.74 & 59.49 & 98.57 & 82.41 \\
\textit{RoLlama3.1-8b-Instruct} & 94.57 & 60.10 & 95.12 & 87.53\\
\hline
gemma-1.1-7b-it & 87.54	&51.49&	83.87	& 85.61 \\
\textit{RoGemma-7b-Instruct} & 86.96	&56.72	&98.80	&85.81\\
\hline
GPT-3.5* & 83.64 & 44.70 & - & - \\
GPT-4o* & 92.78 & 45.22 & - & - \\
\hline
\end{tabular}
\vspace*{-0.6em}
\caption{LaRoSeDa (sentiment analysis) performance. \textbf{Bold} denotes the best result for each setting. *Unlike other models where we framed the problem as a classification task (i.e., comparing which class has the highest logprob), we treat the problem as a generation task for GPT, filtering the generated text. $^{a}$\citet{niculescu2021rogpt2}}
\label{tab:laroseda}
\end{table*}

\begin{table*}[ht]
\centering
\begin{tabular}{lcc|cc}
\hline
\multirow{2}{*}{\textbf{Model}}& \multicolumn{2}{c}{\textbf{Zero/Few shot}} & \multicolumn{2}{c}{\textbf{Fine-tuned}} \\
& EN->RO (\textit{Bleu}) & RO->EN (\textit{Bleu}) & EN->RO (\textit{Bleu}) & RO->EN (\textit{Bleu})\\
\hline
SOTA - mBART$^{a}$& - & - & \textbf{38.50}& 39.90 \\
\hline
Llama-2-7b & 14.90	&26.61&	24.95&	39.09	\\
\textit{RoLlama2-7b-Base} & 10.01&	13.03&	27.85	&39.30\\
\hline
Llama-2-7b-chat & 15.55	&28.53&	19.99&	31.48	\\
\textit{RoLlama2-7b-Instruct} & 27.14	&19.40&	27.63&	39.75\\
\hline
Mistral-7B-Instruct-v0.2 & 18.60 & 33.99 & 26.19 & 39.88 \\
\textit{RoMistral-7b-Instruct} & 28.28	&6.10	&27.70	&40.36\\
\hline
Llama-3-8B-Instruct & 18.89	& 30.98 & 28.02 & 40.28 \\
\textit{RoLlama3-8b-Instruct} & 22.92	&24.28	&27.31	&\textbf{40.52}\\
\hline
Llama-3.1-8B-Instruct & 19.01 & 27.77 & 29.02 & 39.80 \\
\textit{RoLlama3.1-8b-Instruct} & 21.88 & 23.99 & 28.27 & 40.44\\
\hline
gemma-1.1-7b-it & 17.96	& 27.74 & 25.48 & 36.11 \\
\textit{RoGemma-7b-Instruct} & 24.45	&14.20	&25.96	&39.07\\
\hline
GPT-3.5 & 31.83 & \textbf{40.61} & - & - \\
GPT-4o  & \textbf{34.49} & 39.41 & - & - \\
\hline
\end{tabular}
\vspace*{-0.6em}
\caption{WMT'16 (machine translation) performance. \textbf{Bold} as in Table~\ref{tab:laroseda}. $^{a}$\citet{liu-etal-2020-multilingual-denoising}.}
\label{tab:wmt}
\end{table*}

\begin{table*}[ht]
\centering
\begin{tabular}{lcc|cc}
\hline
\multirow{2}{*}{\textbf{Model}}& \multicolumn{2}{c}{\textbf{Zero/Few shot}} & \multicolumn{2}{c}{\textbf{Fine-tuned}} \\
& (\textit{Exact Match}) & (\textit{F1}) & (\textit{Exact Match}) & (\textit{F1})\\
\hline
SOTA - XLM-R$^{a}$  & - & - & \textbf{69.70}& 83.60\\
\hline
Llama-2-7b & 38.91&	56.82&	65.46&	79.42\\
\textit{RoLlama2-7b-Base} & 30.15 &	47.03 &	67.06 &	79.96\\
\hline
Llama-2-7b-chat & 32.35 &	54.00 &	60.34 &	75.98	\\
\textit{RoLlama2-7b-Instruct} & \textbf{45.71}	& \textbf{65.08} &	59.24	&74.25\\
\hline
Mistral-7B-Instruct-v0.2 & 27.92 & 50.71 & 65.46 & 79.73 \\
\textit{RoMistral-7b-Instruct} & 41.09	&63.21	&47.56	&62.69\\
\hline
Llama-3-8B-Instruct & 39.47	& 58.67 & 67.65 & 82.77 \\
\textit{RoLlama3-8b-Instruct} & 18.89	&31.79	&50.84  &65.18\\
\hline
Llama-3.1-8B-Instruct & 44.96 & 64.45 & 69.50 & \textbf{84.31} \\
\textit{RoLlama3.1-8b-Instruct} & 13.59 & 23.56 & 49.41 & 62.93\\
\hline
gemma-1.1-7b-it & 42.10	& 62.30 & 60.34	& 77.40 \\
\textit{RoGemma-7b-Instruct} & 26.03	&41.58	&46.72	&60.79\\
\hline
GPT-3.5 & 35.53	& 60.68 & - & - \\
GPT-4o  &19.31	& 42.89 & - & - \\
\hline
\end{tabular}
\vspace*{-0.6em}
\caption{XQuAD (question answering) performance. \textbf{Bold} as in Table~\ref{tab:laroseda}. $^{a}$\citet{niculescu2021rogpt2}}
\label{tab:xquad}
\end{table*}

\begin{table*}[!pt]
\centering
\begin{tabular}{lcc|cc}
\hline
\multirow{2}{*}{\textbf{Model}}& \multicolumn{2}{c}{\textbf{Few shot}} & \multicolumn{2}{c}{\textbf{Fine-tuned}} \\
& (\textit{Spearman}) & (\textit{Pearson}) & (\textit{Spearman}) & (\textit{Pearson})\\
\hline
SOTA - Romanian BERT$^{a}$ & - & - & 83.15 & 83.70 \\
\hline
Llama-2-7b & 9.08	&9.07	&79.93	&81.08\\
\textit{RoLlama2-7b-Base} & 7.89&	7.98	&71.75	&71.99\\
\hline
Llama-2-7b-chat & 32.56	&31.99	&74.08	&72.64\\
\textit{RoLlama2-7b-Instruct} & 59.69 & 57.16	&84.66	&85.07\\
\hline
Mistral-7B-Instruct-v0.2 & 62.62 & 60.86 & 84.92 & 85.44 \\
\textit{RoMistral-7b-Instruct} & 78.47	& 77.24	& 87.28	&87.88\\
\hline
Llama-3-8B-Instruct & 73.04 & 72.36 & 83.49	& 84.06 \\
\textit{RoLlama3-8b-Instruct} & 77.60 & 76.86 &86.70	&87.09\\
\hline
Llama-3.1-8B-Instruct & 72.11 & 71.64 & 84.59 & 84.96 \\
\textit{RoLlama3.1-8b-Instruct} & 75.89 & 76.00 & 86.86 & 87.05\\
\hline
gemma-1.1-7b-it & 49.10 & 50.23 & 83.43	& 83.64 \\
\textit{RoGemma-7b-Instruct} & 73.23	&71.58	&\textbf{88.42}	&\textbf{88.45}\\
\hline
GPT-3.5 & 79.08	& 78.97 & - & - \\
GPT-4o  &\textbf{84.36}	&\textbf{84.36} & - & - \\
\hline
\end{tabular}
\vspace*{-0.6em}
\caption{Ro-STS (semantic text similarity) performance. \textbf{Bold} as in Table~\ref{tab:laroseda}. $^{a}$~\cite{niculescu2021rogpt2}}.
\label{tab:sts}
\end{table*}

\subsection{RoCulturaBench}
Finally, we were interested in how grounded are LLMs in the historical, cultural, and social realities of Romania. For this reason, we devised a new benchmark, in the form of RoCulturaBench. RoCulturaBench was built manually by a team of Romanian academics from the field of humanities and aims to address all the significant aspects of Romanian culture (in the broad sense of the term, ranging from artistic and scientific contributions to cuisine and sports). The benchmark consists of 100 prompts grouped into 10 domains, covering three types of data (see Appendix Section~\ref{sec:app_roculturabench_example} for examples): (a) communicating information in various formal and informal contexts (the first three categories, concerning grammar and spell check, expressing subjectivity – especially emotions and personal values –, producing functional texts, role-playing and impersonating another person, creativity, and different ways of evaluation and argumentation); (b) factual information about Romania (categories IV to VII, which refer to Romanian geography, tourism, history, demography, politics, science, culture, and sports); (c) information on how Romanians perceive themselves and the world (e.g., traditions, customs, beliefs, stereotypes, representations, attitudes, values).

The design of the benchmark has intentionally avoided any obvious and/or unequivocal answers, such as those resulting from the usage of scientific laws or formulas. The instructions included notions and qualifiers that would each time prompt human subjects to consider at least two possible answers for each prompt (depending on how they understood the notions and qualifiers involved in the prompt) but, at the same time, limit the instructions' field of reference so as to not allow a broad spectrum of possible answers.

We provide reference answers for all prompts and evaluate the LLM-generated answers using GPT-4o\footnote{gpt-4o-2024-08-06} as a judge. Compared to MT-Bench, RoCulturaBench contains only single-turn interactions (i.e., one instruction, one answer).

\section{Results \& Discussions}
\label{sec:results}

In this section, we present and discuss the results of our models on the four evaluations: academic benchmarks, MT-Bench, Romanian downstream tasks, and RoCulturaBench.

\subsection{Academic Benchmarks}
Table~\ref{tab:overview} introduces the academic benchmark results. In the first section, Llama2, and different versions of RoLlama2 foundational model are compared. Increasing the pretraining sequence length boosts the overall performance while adding more pretraining data (up to 20\% of our data) slighty degrades the foundational model's performance. We conjecture that this is due to the quality of the pre-training data used, as CulturaX contains data from different sources, including low-quality sources. Nevertheless, we found foundational models exposed to more training data responding better to instruction fine-tuning, consistently outperforming models trained on less data. For that reason, even if the model has lower performance than some of its counterparts, we perform instruction fine-tuning on RoLlama2-7b-Base that was trained on 20\% of the data with a sequence length of 1024. Still, pre-training on Romanian texts increases the performance on four of the six benchmarks when compared to the Llama2 foundational model. There is a noticeable increase in the Winogrande and HellaSwag benchmarks, arguing for the model's capability to understand Romanian and use it for reasoning. On GSM8k and MMLU, we observe a consistent drop in performance that could be explained by the lack of appropriate data in pre-training, thus making it easier for the model to forget mathematical reasoning (for GSM8k) and knowledge about US history or law. Fine-tuning with instructions further increases the overall score by around 6\% over RoLlama2-base with an increase on all of the six benchmarks.

Turning to another family of models, we notice a significant increase in the performance of the Mistral, Llama3, and Gemma models in Romanian. While we did not find details regarding the languages of the training data for Mistral, Gemma and Llama3 were trained mainly on English texts; as such, their proficiency in understanding Romanian is somewhat surprising. Nevertheless, further exposure to Romanian instructions boosts performance across the board, from 3\% up to 8\%. 

Finally, we evaluate other existing models for Romanian including Okapi-Ro, an LLM specialized in Romanian, and Aya, a multilingual LLM with support for Romanian. As Okapi is a model based on the first Llama model, its lackluster performance is expected. Aya-23 has official support for 23 languages, including Romanian, and it performs on par with the instruct version of RoLlama2 and with Mistral models.

\subsection{MT-Bench}
Performance on MT-Bench is presented in Table \ref{tab:chat-mtbench}. We noticed that even when using a Romanian prompt, most Llama2, Mistral, and Llama3 answers, while somewhat correct, are either entirely in English or have the first few words in Romanian, then reverting to English. We specifically ask the judge to penalize answers that are not in Romanian, but unfortunately, experiments showed that the GPT judge scarcely does this, assigning the same score to English answers irrespective of our request. For this reason, we decided to automatically detect the language of the answers\footnote{\url{www.pypi.org/project/langdetect/}} and penalize the answers that are not written in Romanian with a score of 0. Upon manual inspection, we decided to manually assign the language using simple rules for a few specific prompts where automated methods fail. To alleviate the predisposition of models to answer in English even if the prompt is in Romanian, we added after each instruction an additional request to answer only in Romanian. As seen in the last column in Table \ref{tab:chat-mtbench}, when specifically prompted to answer in Romanian (\textit{ro\_prompted}), the models tended to perform better, with Llama3 and Gemma being capable of answering mostly in the correct language. As expected, this is not an issue with RoLLMs, as they generate text in Romanian without being specifically told to do so.

\begin{table*}[ht]
\centering
\begin{tabular}{lc|c}
\hline
\textbf{Model} & \textbf{Score} & \textbf{\#Answers in Ro}\\
\hline
Llama-2-7b-chat (\textit{ro\_prompted}) & 1.21 & 33 / 100 \\
\textit{RoLlama2-7b-Instruct} & 4.08 & \textbf{100 / 100} \\
\hline
Mistral-7B-Instruct-v0.2 (\textit{ro\_prompted}) & 3.68 & 97 / 100 \\
\textit{RoMistral-7b-Instruct} & 3.99 & \textbf{100 / 100} \\
\hline
Llama-3-8B-Instruct (\textit{ro\_prompted}) & \textbf{4.62} & \textbf{100 / 100} \\
\textit{RoLlama3-8b-Instruct} & 3.81 & \textbf{100 / 100} \\
\hline
Llama-3.1-8B-Instruct (\textit{ro\_prompted}) & 3.54 & \textbf{100 / 100} \\
\textit{RoLlama3.1-8b-Instruct} & \textbf{3.55} & \textbf{100 / 100} \\
\hline
gemma-1.1-7b-it (\textit{ro\_prompted}) & 3.38 & \textbf{100 / 100} \\
\textit{RoGemma-7b-Instruct} & \textbf{3.51} & \textbf{100 / 100} \\
\hline
GPT-3.5 & \textbf{6.09} & \textbf{100 / 100} \\

\hline
\end{tabular}
\vspace*{-0.6em}
\caption{RoCulturaBench performance. \textbf{Bold} denotes as in Table~\ref{tab:overview}.}
\label{tab:roculturabench}
\end{table*}

Turning to the evaluation scores, we notice the gap in performance between the 1st and 2nd turns is larger for RoLLMs than for their counterparts. For RoLLMs, this is somewhat expected as the instruction fine-tuning dataset contains, for the most part, single-turn conversations. Adding more multi-turn conversations in the fine-tuning dataset should also close the performance gap on the second turn, thus leading to greater overall performance. 

We also emphasize that we performed instruction fine-tuning on base models except for the newer Llama3 and Llama3.1 models, for which we lost valuable time and resources spent on alignment. For the Llama3 model family, we found better results by performing direct instruction fine-tuning on the Instruct variant, a model fine-tuned and aligned with human preferences on a huge dataset that also contains 10M human-annotated examples\footnote{https://huggingface.co/meta-llama/Meta-Llama-3-8B-Instruct, last accessed on 1st of October 2024}. To put things into perspective, we translated and used around 2.7M English instructions, out of which only 25k were human-created. As such, we are still investigating how to best train such a model, including continual instruction fine-tuning~\cite{chen2024instructioncp}.

\subsection{Romanian Downstream Tasks}
Downstream tasks performance is presented in Tables~\ref{tab:laroseda},~\ref{tab:wmt},~\ref{tab:xquad} and~\ref{tab:sts}. For sentiment analysis (Table~\ref{tab:laroseda}), we note the strong performance of RoLLMs compared to their counterparts in both settings. The gap in performance shrinks for fine-tuned variants, as all models are somewhat close in performance. It was previously shown that smaller BERT-based models~\cite{avram-etal-2022-distilling,masala2020robert} also achieve comparable performance when fine-tuned~\cite{niculescu2021rogpt2}, especially for the binary case, suggesting that this might be the limit on this particular dataset. 

In the case of machine translation (see Table~\ref{tab:wmt}), we note the discrepancy in translation performance between EN->RO and RO->EN in zero/few-shot setting: RoLLMs perform better in EN->RO setting, while their counterparts are more proficient in RO to EN translations. When finetuned, RoLLMs consistently outperform their counterparts, a trend also observed in the sentiment analysis task. The state-of-the-art model for EN<->RO translations is mBART~\cite{liu-etal-2020-multilingual-denoising}, a sequence-to-sequence denoising auto-encoder that is pre-trained on large-scale English and Romanian dataset and further fine-tuned 
on WMT.

In the case of question answering (see Table~\ref{tab:xquad}), the fine-tuning setup is different due to the nature of the training data as the XQuAD dataset consists of 1190 question-answer pairs professionally translated from the development set of SQuAD v1.1~\cite{rajpurkar-etal-2016-squad}. Therefore, fine-tuning in this case is done on the English dataset, while evaluation is done on the Romanian question-answer pairs. Naturally, non-RO LLMs tend to respond better to data in English. Also, we note that the foundational models perform better than instruct variants in this particular case of XQuAD. In zero/few-shot settings, RoLLMs perform better, with the exception of RoGemma-7b-Instruct and RoLlama3-8b-Instruct, both models exhibiting very strong zero-shot performance and a downward trend especially with 3 and 5 examples. 

Finally, we notice the same pattern in the case of semantic text similarity (see Table~\ref{tab:sts}): strong performance of RoLLMs both when fine-tuned and in a few-shot context. We found in our experiments that most models struggle in the zero-shot setting in this task, generating negative correlation scores. For this reason, we decided to evaluate the semantic text similarity task only on 1, 3, and 5-shot settings, akin to how we evaluated GSM8k. For tasks where the output is numerical and has to adhere to a fixed format, we recommend using at least 1 positive example when using these models.

\subsection{RoCulturaBench}

Results on RoCulturaBench are presented in Table~\ref{tab:roculturabench}. Taking an English foundational model and adapting it to a new language by performing instruction fine-tuning is not enough to properly understand the nuances and realities of a culture. Testament stands the competitive performance of RoLlama2: starting from an arguably weaker model but also pre-training it performs on par with newer and stronger models that have only been fine-tuned.

We argue that, as expected, cultural information and social knowledge are primarily injected into the model in the pre-training phase, not in the fine-tuning phase, especially when fine-tuning exploits non-natively data (i.e., written in English and then translated). As in the case of MT-Bench, Llama3 outperforms other models by a noteworthy margin, with only GPT-3.5 being stronger. For more detailed results and comparisons, see Appendix Section~\ref{sec:app_roculturabench}. In addition, results obtained using the LLM-as-a-judge approach are not the be-all and end-all of evaluations, especially regarding non-English language and culture. We leave human evaluation and other approaches (e.g., training a language-specific judge) for future work.

\begin{table*}[ht]
\centering
\begin{tabular}{lc|c}
\hline
\textbf{Model} & \textbf{RoMT-Bench} & \textbf{RoCulturaBench}\\
\hline
RoLlama2-7b-Instruct & 4.43 & 4.08 \\
RoLlama2-7b-Instruct-DPO & 4.61 & \textbf{4.80} \\
\hline
RoMistral-7b-Instruct & 5.29 & 3.99 \\
RoMistral-7b-Instruct-DPO & 5.88 & 4.72 \\
\hline
RoLlama3-8b-Instruct & 5.38 & 3.81 \\
RoLlama3-8b-Instruct-DPO & 5.87 & 4.40 \\
\hline
RoLlama3.1-8b-Instruct & 5.42 & 3.55 \\
RoLlama3.1-8b-Instruct-DPO & \textbf{6.21} & 4.42 \\
\hline
RoGemma-7b-Instruct & 5.24 & 3.51 \\
RoGemma-7b-Instruct-DPO & 5.48 & 3.94 \\
\hline
\end{tabular}
\vspace*{-0.6em}
\caption{Performance of aligned RoLLMs. \textbf{Bold} denotes the best results for each benchmark.}
\label{tab:aligned}
\end{table*}

\subsection{Human Alignment}

After manually inspecting part of the results for RoMTBench and RoCulturaBench, we found that our models tend to have a more rigid answer. The other models have a more conversational tone, which is more in line with the reference answers and general human preferences. This sometimes rigid tone of RoLLMs stems mostly from the instruction dataset used for fine-tuning and the lack of human preference alignment mechanisms. To test this, we decided to translate the HelpSteer dataset~\cite{wang-etal-2024-helpsteer} containing human preference data and use it to perform Direct Preference Optimization~\cite{rafailov2024direct} (DPO) on our models.

Results are presented in Table~\ref{tab:aligned}. Using only 15k samples, we notice a significant increase in performance across the board, proving the essential role of human alignment for LLMs. We leave a more thorough approach for human alignment, including using more data, to future work.

\section{Conclusions}
\label{sec:conclusion}

This paper introduces the first open-source LLM models specialized for Romanian. The released models show promising results, outperforming existing solutions for a wide array of tasks. 

We present a general training and evaluation recipe which we expect to work on different, more powerful LLMs as well (e.g., models larger than 7-8B, Mixtral\footnote{\url{www.mistral.ai/news/mixtral-of-experts/}} models) as well as for other languages. We firmly believe that this is just the first step towards building and adapting LLMs for Romanian for both research and industry applications.

Pretraining on additional cleaner data, validating existing translations, and adding more instructions, conversations and human preference data are just a few improvements to increase the quality of RoLLMs.

\section*{Limitations}

Our approach for adapting existing LLMs to a new language (i.e., Romanian) has some notable limitations. One significant drawback is the reliance on translated (mainly single-turn) instructions for the instruction fine-tuning step. Subtle nuances might not be captured by the translation process, a process that might introduce inaccuracies that decrease the model's performance. Additionally, our models lack any kind of safeguard or moderation mechanism, meaning that they could generate inappropriate, racist, toxic, dangerous, or even illegal content, raising ethical and practical concerns. 

\section*{Acknowledgements}

Synchronization and collaboration for this work took place within a project\footnote{\url{www.ilds.ro/llm-for-romanian}} carried out in the Institute of Logic and Data Science with funding from BRD Groupe Societe Generale. Part of this work was funded by the EU’s NextGenerationEU instrument through
the National Recovery and Resilience Plan of Romania – Pillar III-C9-I8, managed by the Ministry of Research, Innovation and Digitalization, within the project entitled Measuring Tragedy: Geographical
Diffusion, Comparative Morphology, and Computational Analysis of European Tragic Form (METRA), contract no. 760249/28.12.2023, code CF 163/31.07.2023. We thank Alin Stefanescu and Laurentiu Leustean for participating in various discussions regarding this project.


\bibliography{custom}

\appendix

\section{Detailed results}

\subsection{MT-Bench}
The scores for each category for Romanian MT-Bench are presented in Table~\ref{tab:chat-mtbench-per-categ}. We notice a general upward trend for more scientific categories such as math, extraction (prompts akin to downstream tasks), stem, and with the exception of RoLLama3 even for the writing category. 
On the other hand, our models exhibit degrading performance in coding, reasoning, and roleplaying categories. For the roleplay and code categories, this is expected due to the nature of our instruction dataset (i.e., mostly single-turn conversations and no coding instructions). We believe that a larger, more diverse set of instructions should easily bridge the performance gap in these categories, and even inverse the downward trend.

\begin{table*}[!htb]
\centering
\begin{tabular}{lcccccccc}
\hline
\textbf{Model} & \textbf{W} & \textbf{Rp} & \textbf{R} & \textbf{M} & \textbf{Code} & \textbf{Ext} & \textbf{STEM} & \textbf{H}\\
\hline

Llama-2-7b-chat (\textit{ro\_prompted}) & 1.95 & 1.15 & 0.70 & 1.00 & 0.25 & 1.35 & 1.35 & 0.90\\
\textit{RoLlama2-7b-Instruct-DPO} & 5.60 & 6.25 & 4.35 & 2.30 & 2.25 & 3.95 & 5.35 & 6.80 \\
\hline

Mistral-7B-Instruct-v0.2 (\textit{ro\_prompted}) & 4.70 & 5.00 & 5.40 & 3.80 & 4.85 & 3.65 & 5.50 & 7.30 \\
\textit{RoMistral-7b-Instruct-DPO} & 6.75 & 6.50 & 6.90 & 4.50 & 3.35 & \textbf{5.60} & 5.85 & \textbf{7.60}\\

\hline

Llama-3-8B-Instruct (\textit{ro\_prompted}) & \textbf{7.20} & 6.25 & 7.55 & 4.30 & 5.40 & 4.70 & 5.70 & 6.60\\
\textit{RoLlama3-8b-Instruct-DPO} & 6.90 & 6.60 & 7.25 & 4.80 & 3.35 & 4.55 & \textbf{6.60} & 6.90 \\

\hline

Llama-3.1-8B-Instruct (\textit{ro\_prompted}) & 6.55 & 4.95 & 7.35 & 5.50 & \textbf{5.75} & 3.55 & 5.25 & 6.63\\
\textit{RoLlama3.1-8b-Instruct-DPO} & 6.85 & \textbf{6.65} & \textbf{7.90} & \textbf{6.25} & 3.95 & 4.55 & 6.00 & 7.55 \\

\hline

gemma-1.1-7b-it (\textit{ro\_prompted}) & 5.45 & 5.00 & 5.75 & 3.45 & 3.80 & 4.05 & 5.00 & 6.15\\
\textit{RoGemma-7b-Instruct-DPO} & 5.90 & 5.95 & 6.35 & 5.35 & 3.55 & 4.55 & 5.65 & 6.60\\

\hline


\hline
\end{tabular}
\vspace*{-0.6em}
\caption{Performance on each category of Romanian MT-Bench. W - Writing, Rp - Roleplay, R - Reasoning, M - Math, Code - Coding, Ext - Extraction, H - Humanities. \textbf{Bold} denotes the best result for each category.}
\label{tab:chat-mtbench-per-categ}
\end{table*}

\subsection{RoCulturaBench}
\label{sec:app_roculturabench}

\begin{table*}[ht]
\centering
\begin{tabular}{lccc}
\hline
\textbf{Model} & \textbf{W/ Ref} & \textbf{W/o Ref} & \textbf{Score diff}\\
\hline
Llama-2-7b-chat (\textit{ro\_prompted}) & 1.21  & 2.42 & 1.21\\
\textit{RoLlama2-7b-Instruct-DPO} & 4.80 & 5.95 & 1.15\\
\hline

Mistral-7B-Instruct-v0.2 (\textit{ro\_prompted}) & 3.68 & 4.48 & 0.80\\
\textit{RoMistral-7b-Instruct-DPO} & 4.72 & 5.88 & 1.16 \\

\hline

Llama-3-8B-Instruct (\textit{ro\_prompted}) & 4.62 & 5.60 & 0.98 \\
\textit{RoLlama3-8b-Instruct-DPO} & 4.40 & 5.38 & 0.98\\

\hline

Llama-3.1-8B-Instruct (\textit{ro\_prompted}) & 3.54 & 4.50 & 0.96 \\
\textit{RoLlama3.1-8b-Instruct-DPO} & 4.42 & 5.51 & 1.09\\

\hline

gemma-1.1-7b-it (\textit{ro\_prompted}) & 3.38 & 4.07 & 0.69\\
\textit{RoGemma-7b-Instruct-DPO} & 3.94 & 4.71 & 0.77\\
\hline


\hline
\end{tabular}
\vspace*{-0.6em}
\caption{RoCulturaBench performance with and without reference answers.}
\label{tab:roculturabench-ref}
\end{table*}

In Table~\ref{tab:roculturabench-ref}, we present results on RoCulturaBench with and without taking into account the human-written reference answers. First, we notice an overall lower score when the judge is given the reference answer and is asked to compare it to the output of a model. This decrease in performance is consistent and easily noticeable for every model. But this decrease is not completely uniform: for RoLLMs, the penalty tends to be lower (e.g., in the case of Llama3, we notice a loss of 1.53, while for RoLLama3, the score is lower by 1.23).

Scores for each category of RoCulturaBench are presented in Table~\ref{tab:roculturabench-per-categ}. We note the very strong performance of GPT-3.5 of Llama3 across all categories. The Role category seems to be the closest category in terms of scores, while for the Geography and Sports categories, both GPT-3.5 and Llama3 really shine compared to all other models.

\begin{table*}[!hbt]
\centering
\begin{tabular}{lcccccccccc}
\hline
\textbf{Model} & \textbf{W} & \textbf{R} & \textbf{A} & \textbf{G} & \textbf{H} & \textbf{C\&S} & \textbf{Sp} & \textbf{C\&T} & \textbf{S} & \textbf{C}\\
\hline

Llama-2-7b-chat (\textit{ro\_prompted}) & 0.0 & 1.5 & 1.5 & 1.6 & 2.2 & 1.7 & 0.3 & 0.6 & 1.4 & 1.3\\
\textit{RoLlama2-7b-Instruct-DPO} & 5.4 & 6.0 & 6.0 & 5.2 & 4.8 & 3.9 & 3.4 & 4.7 & 4.3 & 4.3\\
\hline

Mistral-7B-Instruct-v0.2 (\textit{ro\_prompted}) & 4.1 & 3.3 & 5.6 & 3.6 & 3.7 & 3.5 & 3.1 & 2.7 & 4.5 & 2.7\\
\textit{RoMistral-7b-Instruct-DPO} & 5.6 & 5.7 & 7.1 & 5.5 & 3.6 & 2.9 & 4.3 & 4.1 & 4.9 & 3.5\\

\hline

Llama-3-8B-Instruct (\textit{ro\_prompted}) & 5.4 & 5.8 & 6.7 & 4.9 & 3.8 & 3.5 & 3.3 & 4.2 & 5.0 & 3.6\\
\textit{RoLlama3-8b-Instruct-DPO} & 5.3 & 5.6 & 5.8 & 5.5 & 3.7 & 2.9 & 3.8 & 4.0 & 4.6 & 2.8 \\

\hline
Llama-3.1-8B-Instruct (\textit{ro\_prompted}) & 4.1 & 4.3 & 5.3 & 3.6 & 2.7 & 2.6 & 3.0 & 3.1 & 4.0 & 2.7\\
\textit{RoLlama3.1-8b-Instruct-DPO} & 4.9 & 5.7 & 6.8 & 5.2 & 3.5 & 3.4 & 3.1 & 3.7 & 4.7 & 3.2 \\

\hline

gemma-1.1-7b-it (\textit{ro\_prompted}) & 3.9 & 4.8 & 5.6 & 3.6 & 3.2 & 2.3 & 2.5 & 2.7 & 3.4 & 1.8 \\
\textit{RoGemma-7b-Instruct-DPO} & 5.0 & 5.0 & 6.5 & 3.8 & 3.4 & 2.2 & 3.0 & 3.5 & 4.9 & 2.1\\
\hline
Gpt-3.5 & \textbf{6.0} & \textbf{7.0} & \textbf{7.8} & \textbf{7.1} & \textbf{5.7} & \textbf{4.8} & \textbf{4.9} & \textbf{6.6} & \textbf{5.8} & \textbf{5.2}\\

\hline


\hline
\end{tabular}
\vspace*{-0.6em}
\caption{Performance on each category of RoCulturaBench. W - Writing, R - Roles, A - Argumentation, G - Geography, H - History, C\&S - Culture \& Science, C\&T - Customs \& Traditions, Sp - Sports, S - Stereotypes, C - Celebrity. \textbf{Bold} denotes the best result for each category.}
\label{tab:roculturabench-per-categ}
\end{table*}

\section{RoCulturaBench Example}
\label{sec:app_roculturabench_example}

In Table~\ref{tab:prompts-robench} we present a few examples prompts from each category of RoCulturaBench in both Romanian and English.

\begin{table*}[!hbt]
\centering
\begin{tabular}{cp{5.6cm}p{5.6cm}}
\hline
\textbf{Category} & \textbf{Romanian Prompt} & \textbf{English Prompt}\\
\hline\\[-0.3cm]
\multirow{12}{*}{Writing} & Redactează, în calitate de angajat, o cerere în care să-i soliciți într-un mod convingător managerului filialei locale a companiei la care lucrezi să-ți aprobe organizarea unei petreceri cu ocazia zilei de 8 Martie. Precizează că petrecerea se va desfășura în Sala de Conferințe, va începe cu două ore înainte de terminarea programului, va include 50 de participanți și va avea nevoie de un buget de 5.000 de lei din fondul de protocol al filialei. & Write, as an employee, a request in which you convincingly ask the manager of the local branch of the company you work for to approve your organization of a party on the occasion of March 8. It states that the party will take place in the Conference Hall, will start two hours before the end of the program, will include 50 participants and will need a budget of 5,000 lei from the protocol fund of the branch. \\[5.35cm]

\hline\\[-0.3cm]

\multirow{3}{*}{Role} & Imaginează-ți că ești o pisică vagaboandă în București. Povestește-ne cea mai fericită zi din viața ta.
& Imagine you are a stray cat in Bucharest. Tell us about the happiest day of your life. \\[1.05cm]

\hline\\[-0.3cm]

\multirow{8.3}{*}{Argumentation} & Scrie un text de maximum 300 de cuvinte în care să argumentezi pentru drepturile României asupra Transilvaniei. & Write a text of a maximum of 300 words in which you argue for Romania's rights over Transylvania. \\[1.6cm]
 & Explică și argumentează în maxim 300 de cuvinte afirmația lui Lucian Blaga că „veşnicia s-a născut la sat”. & Explain and argue in maximum 300 words Lucian Blaga's statement that "eternity was born in the village". \\[1.1cm]

\hline\\[-0.3cm]

\multirow{3}{*}{Geography} & Ce oraș medieval din România îmi recomanzi să vizitez? Descrie-mi-l în maxim 200 de cuvinte. & Which medieval city in Romania do you recommend I visit? Describe it to me in no more than 200 words. \\[1.05cm]

\hline\\[-0.3cm]

\multirow{3}{*}{History} & Cum a evoluat popularitatea lui Nicolae Ceaușescu de-a lungul perioadei în care a condus România? & How did the popularity of Nicolae Ceaușescu evolve during the period in which he ruled Romania? \\[1.05cm]

\hline\\[-0.3cm]

\multirow{5.25}{*}{Culture \& Science} & Ce înseamnă „Noul Val” din cinematografia românească? & What does "New Wave" mean in Romanian cinema? \\[0.675cm]

& Ce contribuții au avut românii în domeniul aviației? Oferă-mi trei exemple. & What contributions did the Romanians have in the field of aviation? Give me three examples. \\[1.05cm]

\hline\\[-0.3cm]
 
\multirow{2}{*}{Sport} & Cât de popular este tenisul în România? & How popular is tennis in Romania? \\[0.55cm]

\hline\\[-0.3cm]

\multirow{3}{*}{Customs \& Tradition} & Ce sunt horele și cât de actuale mai sunt astăzi aceste obiceiuri în România? & What are "hora" dances and how current are these customs in Romania today? \\[1cm]

\hline\\[-0.3cm]

\multirow{3}{*}{Stereotypes} & În ce țări preferă românii să emigreze? Detaliază cele mai importante cinci exemple. & To which countries do Romanians prefer to emigrate? Detail the five most important examples. \\[1.05cm]

\hline\\[-0.3cm]

\multirow{2}{*}{Celebrity} & Care este cel mai cunoscut scriitor român pe plan internațional? & Who is the best-known Romanian writer internationally? \\[0.55cm]

\hline
\end{tabular}
\vspace*{-0.6em}
\caption{Sample prompts from RoCulturaBench}
\label{tab:prompts-robench}
\end{table*}

\section{Prompts for Downstream Tasks}
\label{sec:app_prompts}

The prompts used for the for downstream tasks are available in \autoref{tab:prompts-downstream}. The models received only the Romanian prompts, however we also included their translations into English for increased readability.

\begin{table*}[!hbp]
\centering
\begin{tabular}{ccp{5.4cm}p{5.4cm}}
\hline
\textbf{Dataset} & \textbf{Scenario} & \textbf{Romanian Prompt} & \textbf{English Prompt}\\
\hline\\[-0.3cm]
\multirow{11}{*}{LaRoSeDa} & \multirow{5}{*}{Binary} & \makecell[lt]{Analizează următoarea recenzie și\\evalueaz-o ca fiind pozitivă sau\\negativă.\\Recenzie: \{\textit{review}\}\\Evaluare:} & \makecell[lt]{Analyze the following review\\and evaluate it as positive or\\negative.\\Review: \{\textit{review}\}\\Evaluation:} \\[2.15cm]
& \multirow{6}{*}{Multi-Class} & \makecell[lt]{Analizează următoarea recenzie și\\caracterizează nota oferită produsu-\\lui pe o scară de la 1 la 5, cu urmă-\\toarele opțiuni: 1, 2, 4 sau 5.\\Recenzie: \{\textit{review}\}\\Notă:} & \makecell[lt]{Analyze the following review and\\rate the product on a scale of 1 to 5\\with the following options: 1, 2, 4\\or 5.\\Review: \{\textit{review}\}\\Rating:} \\
\hline\\[-0.3cm]
\multirow{4}{*}{WTM16} & \multirow{2}{*}{En-Ro} & Tradu următorul text din engleză în română: \{\textit{text}\}& Translate the following text from English to Romanian: \{\textit{text}\}\\[0.7cm]
& \multirow{2}{*}{Ro-En} & Tradu următorul text din română în engleză: \{\textit{text}\}&  Translate the following text from Romanian to English: \{\textit{text}\}\\[0.6cm]

\hline\\[-0.3cm]

\multirow{3}{*}{XQuAD} & \multirow{3}{*}{-} & \makecell[lt]{\{\textit{Context}\}\\Întrebare: \{\textit{Question}\}\\Răspuns:} & \makecell[lt]{\{\textit{Context}\}\\Question: \{\textit{Question}\}\\Answer:} \\[1.1cm]

\hline\\[-0.3cm]

\multirow{7}{*}{RoSTS} & \multirow{7}{*}{-} & \makecell[lt]{Generează un număr între 0 și 1\\care descrie similaritatea semantică\\dintre următoarele două propoziții:\\Propoziție 1: \{\textit{Sentence}\}\\Propoziție 2: \{\textit{Sentence}\}\\Scor de similaritate semantică:} & \makecell[lt]{Generate a number between 0 and\\1 that describes the semantic sim-\\ilarity between the following two\\sentences:\\Sentence 1: \{\textit{Sentence}\}\\Sentence 2: \{\textit{Sentence}\}\\Semantic Similarity Score:} \\[3cm]

\hline
\end{tabular}
\vspace*{-0.6em}
\caption{Prompts used for downstream tasks and corresponding English variant}
\label{tab:prompts-downstream}
\end{table*}

\section{Model Architecture}
As we continue the process of pre-training or instruction fine-tuning on the foundational model, for each RoLLM we inherit the architecture and the vocabulary of the English variant. Specifically, RoLlama2-7b models use the standard Transformer~\cite{vaswani2017attention} architecture with 32 attention heads, 32 layers, hidden layers of size 4096, and a vocabulary of size 32000. Mistral shares, for the most part, the same architecture with Llama2, adding a Sliding Window Attention~\cite{beltagy2020longformer} mechanism, keeping the same vocabulary. Llama3-8B and, by extension RoLlama3-8b-Instruct, adds grouped-query attention~\cite{ainslie-etal-2023-gqa} on top of the Llama2 architecture and also increases the vocabulary size up to 128k. Finally, RoGemma-7b uses 28 layers, 16 attention heads, hidden layers of size 3072 and a vocabulary size of 256k tokens.

\section{Computational Resources}

Pre-training was performed on up to six NVIDIA A100s with 80GB of memory. Continual pre-training on 5\% of the data using a sequence length of 512 for one epoch used 320 GPU hours, 468 GPU hours were consumed for a sequence length of 1024, while for a sequence length of 2048, 708 GPU hours were needed. Instruction fine-tuning of RoLlama2 with a sequence length of 1024 for one epoch required 120 GPU hours, while performing DPO required 30 GPU minutes for one epoch. Evaluations were performed on a single A100 GPU at a time.
\end{document}